\documentclass[11pt]{article}
\usepackage{acl}

\usepackage{times}
\usepackage{latexsym}
\usepackage[T1]{fontenc}
\usepackage[utf8]{inputenc}
\usepackage{amsmath,amssymb}
\usepackage{booktabs}
\usepackage{graphicx}
\usepackage{microtype}
\usepackage{orcidlink}
\graphicspath{{figures/}}

\title{Estimating Silenced Tokens in Quantized LLMs}

\author{Giuseppe Pappalardo\,\orcidlink{0000-0001-6938-5234} \\
  University of Enna ``Kore'', Italy \\
  \texttt{giuseppe.pappalardo@unikore.it}}

\begin{document}
\maketitle

\begin{abstract}
Quantization is the standard way to shrink large language models, and it
is usually judged by perplexity and task accuracy. We ask a different
question: after quantization, which words can the model still say at all?
A token is \emph{silenced} when no hidden state can make it the top choice
of the output layer, so a greedy decoder can never produce it. We give a
simple, scalable way to estimate how many tokens a quantized output head
silences, combining three checks: a direction that lets a token win
(so it is safe), an upper bound on how close a token's row is to being
trapped inside the others, and an exact match that proves a token can never
win. Each count is therefore a certified range: exact matches give its
lower end and flagged tokens can only raise its upper end. In nearly
all settings we report down to 3 bits the two ends coincide, so the counts
are exact. Across three open models we find that quantization silences tokens: on
large-vocabulary models it already appears at 8-bit precision---up to a few
hundred previously reachable tokens on a 1B-parameter model---and grows as
precision drops, while a small-vocabulary model resists until 3 bits. The
loss is far from uniform across languages: on a multilingual vocabulary the silenced tokens
are overwhelmingly non-Latin (Korean, Chinese, Thai, Arabic and Hebrew),
even though Latin tokens dominate the vocabulary. Silencing is thus a
measurable, language-sensitive side effect that standard averages miss,
and it should be reported alongside them.\footnote{Code and per-token
  measures: \url{https://github.com/gpappalardo/EACL2027Paper}}
\end{abstract}

\section{Introduction}

Post-training quantization is the default way to fit a large language model
(LLM) into a tight memory and latency budget \citep{gholami2021survey}.
Its cost is almost always summarized by an \emph{average}: a perplexity
gap, a drop in benchmark accuracy, or a weight-space error. These averages, however, are
blind to an all-or-nothing failure: after quantization some tokens simply
can no longer be said. No matter what the network computes, they never win
the output layer, so a greedy decoder can never emit them. We call such
tokens \emph{unargmaxable} \citep{grivas2022}, or \emph{silenced}.

Whether a token can win the output layer depends only on the geometry of
the output head---the final weight matrix that turns a hidden state into
one score per token. Quantization changes that matrix, and with it the set
of tokens that can still win. Our goal is to \emph{estimate} how large the
silenced set becomes, and which tokens fall into it.
Silencing is a structural property: a silenced token cannot be output for
any input. This differs from a token that simply does not appear in the
output on a given corpus, which can happen for many reasons; we keep the
two notions separate throughout.

Measuring this exactly over a full vocabulary is expensive, so we estimate
it with cheap, one-sided checks (Section~\ref{sec:method}): a token is
declared safe as soon as we find a hidden state that makes it win; a token
is flagged when its row is provably close to being trapped inside the
others; and it is proven silenced when its row exactly coincides with a
combination of the rest. This yields bounds: tokens proven silenced by exact collision give the
lower bound, and flagged tokens can only add to the upper bound. Down to
3 bits almost every silenced token is an exact collision, so the two
bounds essentially coincide. The flagged tokens that remain sit in
degenerate clusters of under-trained rows or in collapsed 2-bit heads,
where resolving them exactly would cost much more computation for little
insight.

We apply the method to three open models---GPT-2, Llama-3.2-1B and
Qwen2.5-0.5B---under standard round-to-nearest (RTN) quantization. Our
findings are:

\begin{itemize}
\item \textbf{Silencing appears early.} On Llama-3.2-1B, whose
full-precision head silences no token, 8-bit quantization already silences
between 160 and 284 tokens depending on the scheme, and the count grows at
lower precision (Section~\ref{sec:silence}).
\item \textbf{Silencing is structurally selective.} The silenced tokens are not random: reserved and under-trained ``glitch'' tokens, numeric tokens, and whole scripts are disproportionately impacted (Section~\ref{sec:silence}).
\item \textbf{It is language-skewed.} On the multilingual Qwen vocabulary
the silenced tokens are overwhelmingly non-Latin---Korean, Chinese, Thai,
Arabic and Hebrew---even though Latin tokens dominate the vocabulary,
whereas the two English-centric models lose mostly Latin tokens
(Section~\ref{sec:lang}).
\end{itemize}

Silencing does not always translate into a measurable quality loss, but it
can, and it does so unevenly across tokens and languages. It is a concrete,
checkable property of a quantized head that complements perplexity and
accuracy.

The practical stakes are highest for multilingual deployment, where 4-bit
weights are routine and the served languages are exactly the ones our
measurements show to be most affected. Quantized checkpoints are widely used, and more so
for larger models (Table~\ref{tab:downloads}): in the Qwen2.5 family the
quantized share of downloads grows from 8\% to 20\% with model size, while
it is almost zero for GPT-2. About 40\% of these quantized downloads are
GGUF files, a format that quantizes the output head. A model can look healthy on an
English benchmark while quietly losing the ability to emit certain Korean
or Thai tokens. Reporting a single average hides this; a silenced-token
count, broken down by script, surfaces it at almost no cost.

\begin{table}[t]
\centering\small
\setlength{\tabcolsep}{3pt}
\begin{tabular}{lrrrrr}
\toprule
Model & Params & GB & Original & Quant. & Share \\
\midrule
GPT-2        & 0.12B &  0.55 & 982.5M &  0.3M & 0.03\% \\
Qwen2.5-0.5B & 0.49B &  0.99 & 112.1M & 10.1M & 8\% \\
Llama-3.2-1B & 1.24B &  2.47 & 170.9M & 55.9M & 25\% \\
Qwen2.5-1.5B & 1.54B &  3.09 & 245.9M & 33.7M & 12\% \\
Qwen2.5-7B   & 7.62B & 15.23 & 219.1M & 56.1M & 20\% \\
\bottomrule
\end{tabular}
\caption{Downloads on the Hugging Face Hub since release (as of November
2026), base and Instruct variants together, of the original models and of
their quantized derivatives. GB is the size of the original weights (GPT-2
in fp32, the others in bf16). Counts of quantized downloads are lower
bounds: they include only repositories that declare the base model.}
\label{tab:downloads}
\end{table}

\section{Background}\label{sec:bg}

\begin{figure}[t]
  \centering
  \includegraphics[width=0.72\linewidth]{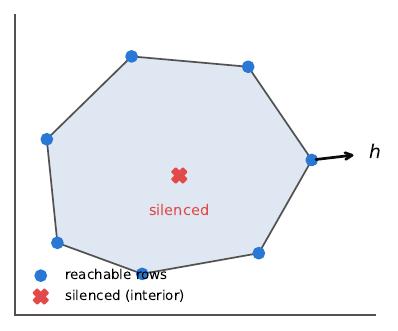}
  \caption{A token can win only if its row lies on the boundary of the set
  of rows: a direction $h$ then pushes it strictly ahead. A row in the
  interior (red) is silenced---no direction makes it the strict top score.
  Quantization moves rows onto a grid and can push a boundary row inside.}
  \label{fig:hull}
\end{figure}

Write the output head as a matrix $W\in\mathbb{R}^{V\times d}$ with one row
$w_t$ per token, acting on a final hidden state $h\in\mathbb{R}^d$ to
produce scores $z=Wh$. Token $t$ is \emph{argmaxable} if some hidden state
makes it the strict winner, i.e.\ $w_t^\top h > w_j^\top h$ for all
$j\neq t$; otherwise it is \emph{silenced}. A classical fact makes this
geometric: $t$ is silenced exactly when its row $w_t$ lies inside the
convex hull of the other rows \citep{grivas2022}. Intuitively, if a row is
surrounded by the others, every direction that raises its score raises some
competitor's score at least as much (Figure~\ref{fig:hull}).

This is the structural counterpart of the ``stolen probability'' effect,
where rare tokens are crowded out of the probability simplex
\citep{demeter2020}, and it is tied to the low-rank softmax bottleneck
\citep{yang2018softmax} that also lets one recover hidden geometry through
an API \citep{finlayson2024}. Prior work studies this geometry in
full-precision models \citep{grivas2022,demeter2020,yang2018softmax}. We ask how quantization reshapes it.

Because argmaxability depends on $W$ alone, analysing the head in isolation
is exact for this question: a token that no hidden state can push to a win
is unreachable for \emph{any} input the rest of the network could produce.
The converse does not hold: a token we call safe has a hidden state that
makes it win, but the model may never produce that hidden state for any
real input. So a safe token is not guaranteed to be reachable in
practice, while a silenced token is unreachable for sure under greedy
decoding. Our silenced counts are therefore conservative: the true number
of tokens the model can never output can only be larger.

\section{Estimating the silenced set}\label{sec:method}

Enumerating the silenced tokens exactly means solving one geometry problem per token over a vocabulary of tens or hundreds of thousands of rows, which does not scale. We instead estimate the set with three one-sided checks that are computationally lightweight on a single consumer-grade GPU, each carrying explicit geometric semantics.

\paragraph{A winning direction (safe).} 
For the vast majority of tokens, we identify a hidden state $h$ that makes the token the strict winner of the output layer. We store this state as a \emph{winning direction}: verifying its validity requires only a single matrix multiplication, a direct numerical check that the token is not silenced. We require strict inequalities throughout our checks; accepting ties would incorrectly classify boundary cases, such as digit tokens, as reachable. As an example, consider two tokens with the same
row, $w_a = w_b = (1,0)$. Every hidden state gives them the same score, so
neither can ever be the unique top choice: they can only tie for it. With
non-strict inequalities both would count as reachable; with strict ones
both are silenced. A decoder that breaks ties by token index could still
emit the lower-index one, but that outcome is fixed by the tie rule, not by
the model, so we count both as silenced. Quantization creates exactly this
case for GPT-2's digit rows at 3 bits. In our counts such ties concern at most one token per group of identical
rows: at most 42 of the 175 tokens GPT-2 silences at 3 bits, 33 of Llama's
264 and 5 of Qwen's 834 at 8 bits.    

\paragraph{A trapping bound (flagged).} When no winning direction is found,
we try to write the token's row as a weighted average of the other rows. Let
$d_t$ be the distance from $w_t$ to the convex hull of the others. This
distance controls the best margin the token can ever achieve:
\begin{equation}
d_t \;=\; \max_{\lVert h\rVert\le 1}\ \min_{j\ne t}\ (w_t-w_j)^\top h ,
\label{eq:duality}
\end{equation}
so an upper bound $d_t\le u$ caps the best achievable winning margin by
$u\,\lVert h\rVert$ for every hidden state $h$. A small $d_t$ therefore
means the token can win, if at all, only by a vanishing margin. We report
such tokens as \emph{flagged}---close to being trapped---together with the
bound. This is an upper bound on $d_t$, not a proof that the token is
silenced, which is why flagged tokens sit at the upper, uncertain end of
our estimate.

\paragraph{An exact match (proven silenced).} In the strongest case the
token's row coincides exactly with a combination of the others---in
practice, bit-for-bit with a single other row. Then the token is provably
silenced: no strict win exists. Quantization creates such exact
collisions when it maps distinct rows onto the same grid point.

The output is therefore a range: a lower end from tokens proven silenced by
exact collision, and an upper end that also includes flagged tokens whose
status is not certain. We report both, which is why we speak of estimating
the silenced set rather than counting it.

\paragraph{Computational pipeline and scalability.} Our estimation framework operates as a data-free, two-stage pipeline executed entirely via batched GPU matrix operations on the stored weight matrix. In the first stage, we evaluate a curated set of candidate directions; identifying a strictly positive margin immediately classifies the token as safe, rapidly resolving most of the vocabulary. For unresolved tokens, we run a first-order projection to approximate $w_t$ as a convex combination of surrounding rows, thereby tightening the upper bound on $d_t$. If the residual converges to zero, we test for a bit-for-bit exact collision. Because the computational footprint is dominated by standard matrix multiplications, evaluating a full $152$k-token vocabulary across a specific model, bit-width, and quantization scheme completes in minutes to a couple of hours on a single consumer GPU. Code and per-token measures are available at \url{https://github.com/gpappalardo/EACL2027Paper}.

\paragraph{Stability on degenerate row clusters.} Because our geometric checks evaluate static weight matrices directly, a binary classification into safe or flagged tokens can become ill-conditioned when rows are densely clustered. For example, Qwen2.5-0.5B contains a degenerate cluster of roughly $1{,}400$ under-trained rows with nearest-neighbour distances of $10^{-5}$ to $10^{-4}$. At such scales, a standard rounding step is comparable to the inter-row separation, making binary reachability classifications intrinsically unstable under quantization. To address these dense regimes, we report the continuous margin bounds from Eq.~\eqref{eq:duality} rather than binary classifications, and we consistently evaluate quantized heads against a shared full-precision baseline to ensure a controlled differential comparison.

\section{Setup}\label{sec:setup}

\paragraph{Models.} 
We study GPT-2 (124M, 50k vocabulary), Llama-3.2-1B
(128k vocabulary) and Qwen2.5-0.5B (152k multilingual vocabulary), spanning
small vs.\ large and English-centric vs.\ multilingual vocabularies.
To check whether the results hold across model size, we also use
Qwen2.5-1.5B and Qwen2.5-7B, which share Qwen2.5-0.5B's vocabulary, and
Llama-3.1-70B, which shares Llama-3.2-1B's (Sections~\ref{sec:silence}
and~\ref{sec:lang}).

\paragraph{Quantization.} We use round-to-nearest (RTN) fake quantization
of all linear weights (attention and MLP projections and the output head;
embeddings, normalization and biases are left in full precision), at
$\{8,4,3,2\}$ bits, symmetric or asymmetric, per-tensor or per-channel.
RTN is the simplest and most widely used scheme and is a clean stress test;
stronger quantizers \citep{frantar2023gptq,lin2024awq,dettmers2022llmint8,
dettmers2023qlora} start from the same rounding step.

\paragraph{What we measure.} For every model, scheme and bitwidth we run the
estimator of Section~\ref{sec:method} on the output head and count the
tokens that become silenced relative to full precision. We treat symmetric
per-channel RTN as the controlled stress test (a single scale per row),
asymmetric per-channel as the closer proxy for realistic deployment, and we
report per-tensor variants as well because they behave differently: a
shared per-tensor scale rounds every row against the same grid, which can
either protect or expose tokens depending on how the rows are distributed.
For reference we also record perplexity (WikiText-2) and ARC-Easy accuracy
so that silencing can be read next to standard quality numbers.

\section{Quantization silences tokens}\label{sec:silence}

\begin{figure*}[t]
  \centering
  \includegraphics[width=\textwidth]{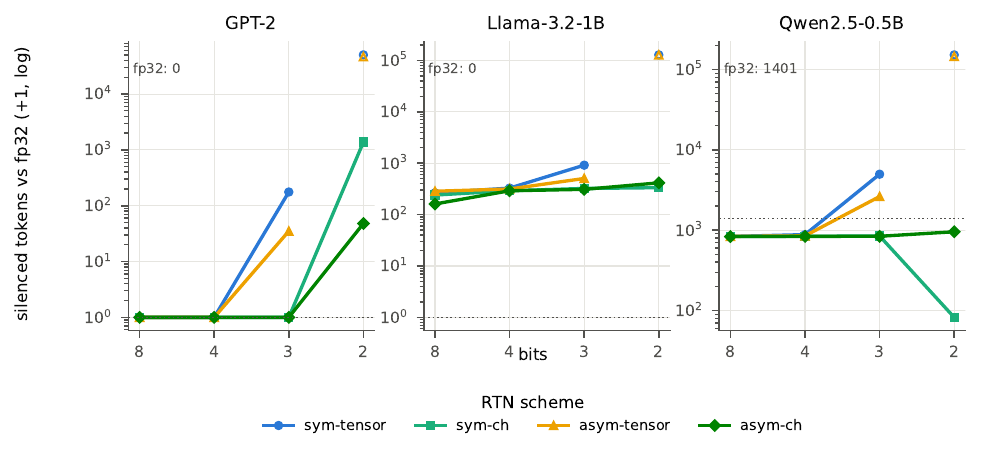}
  \caption{Tokens silenced by quantization relative to full precision
  ($+1$, log scale), by bitwidth and RTN scheme. Full-precision counts are
  shown as dotted lines. Open markers are upper-bound-only cells (the
  vocabulary has largely collapsed). Llama-3.2-1B starts from zero and is
  silenced already at 8 bits.}
  \label{fig:bits}
\end{figure*}

Figure~\ref{fig:bits} shows, for each model, how many tokens become
silenced relative to full precision as precision drops. Two regimes stand
out.

\paragraph{Large vocabularies silence tokens already at 8 bits.}
Llama-3.2-1B is the cleanest case: in full precision its head silences no
token, yet every 8-bit scheme silences between 160 and 284 tokens, rising
through 4 bits and below.
The effect does not fade with scale: Llama-3.1-70B, which shares
Llama-3.2-1B's vocabulary, also silences no token in full precision and
221--287 tokens at 8 bits, all exact collisions and again mostly reserved
special tokens (254 of 287). Qwen2.5-0.5B, whose full-precision head already
has a large cluster of near-duplicate under-trained rows, silences a
further 830 to 834 tokens at 8 bits. The effect is present before any
visible quality loss: at 8 bits both models stay within three accuracy
points of full precision (Table~\ref{tab:main}).

\paragraph{Small vocabularies resist longer, then break.} GPT-2 silences
no token down to 4 bits, then at 3-bit per-tensor silences 175 tokens,
all \emph{proven} silenced by exact collision: the 3-bit grid maps their
rows onto identical vectors. Crucially, 70 of them are numeric tokens
(single digits, years, round numbers).

\begin{figure*}[t]
  \centering
  \includegraphics[width=\textwidth]{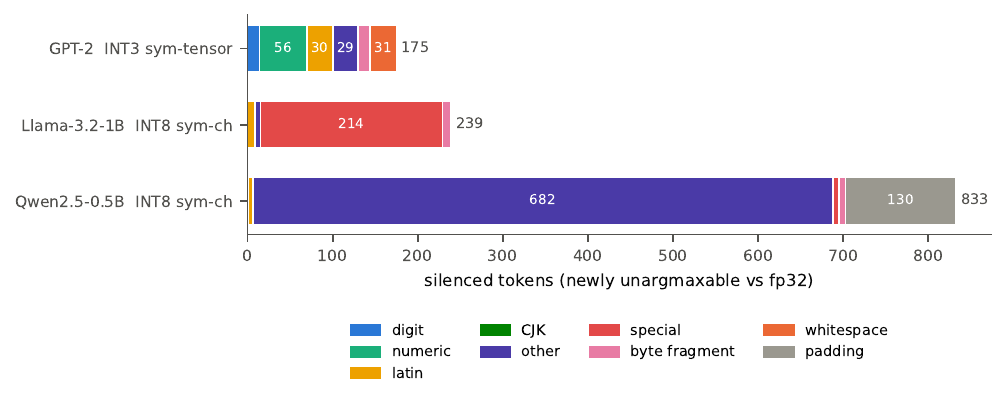}
  \caption{Composition of the silenced sets for one configuration per
  model (GPT-2 at 3 bits, Llama and Qwen at 8 bits). GPT-2 loses digits and numeric tokens; Llama
  loses reserved/special tokens; Qwen loses mostly non-Latin-script tokens
  (``other'').}
  \label{fig:cats}
\end{figure*}

\paragraph{Which tokens fall silent.} The silenced sets are structured
(Figure~\ref{fig:cats}). On Llama they are dominated by reserved special
tokens and known under-trained ``glitch'' tokens; on GPT-2 at 3 bits by the
numeric vocabulary; on Qwen by non-Latin-script tokens. The composition
matters for whether silencing does any harm: losing a deep-tail reserved
token is harmless, losing digits is not.

\paragraph{Group-wise scaling does not prevent silencing.} It is natural to ask whether increasing scale granularity avoids structural silencing. Our measurements show it does not: repeating the sweep with per-group scales along the rows (groups of 32 and 128, the granularity of popular k-quant formats) leaves the pattern essentially unchanged---at 4 bits with groups of 32, Llama-3.2-1B still silences $258$--$289$ tokens (versus $290$ for plain per-channel) and Qwen2.5-0.5B reaches $2232$--$2234$ in total (versus $2236$--$2238$). Silencing is a property of rounding rows onto a shared grid, not an artifact of coarse scaling.

\paragraph{Calibrated quantizers.} We repeated the measurement with GPTQ
\citep{frantar2023gptq} and AWQ \citep{lin2024awq} on Llama-3.2-1B and
Qwen2.5-0.5B, 1.5B and 7B. AWQ keeps the output head in 16 bits and
introduces no new silenced token relative to full precision. Against an
fp16 copy of the head, Qwen2.5-0.5B shows 75 new silenced tokens, but all
75 come from the cast to fp16, not from AWQ quantization itself. GPTQ, as implemented in GPTQModel,
quantizes the head to 8 bits with groups of 32 by default. With this head
Qwen2.5-0.5B silences 180 new tokens (134 when the rest of the model is at
4 bits), all exact collisions, while Llama-3.2-1B and Qwen2.5-1.5B silence
none. When the head is quantized to 4, 3 or 2 bits, GPTQ silences 91, 135
and 259 new tokens on Llama-3.2-1B and 796, 817 and 834 on Qwen2.5-0.5B,
again all exact collisions. On Qwen2.5-7B, GPTQ applied to the head alone
silences 281--285 new tokens, flagged rather than proven. Calibration thus
reduces silencing at the default 8-bit head, but does not prevent it once
the head is quantized to fewer bits.

\paragraph{A note on deployment quantizers.} Production kernels such as
8-bit and NF4 \citep{dettmers2022llmint8,dettmers2023qlora} often leave the
output head in higher precision, so they do not silence tokens
structurally. They can still change decisions: on a 500-document corpus (WikiText-103),
bitsandbytes 4-bit quantization (its default FP4 format) flips $29.4\%$ of
Qwen2.5-0.5B's greedy choices, and $1842$ token types reached in half
precision (fp16) are not reached under it ($1793$ of them Latin).
This is a corpus-level observation, not structural silencing, and we keep
the two notions separate throughout: our counts concern what the head can
\emph{ever} produce, not what it happens to produce on a given corpus.

\paragraph{Silencing is a separate axis from average quality.}
Table~\ref{tab:main} shows that the two do not move together. GPT-2 at 4
bits silences \emph{no} token yet its perplexity has already exploded by
orders of magnitude: average quality can collapse with the reachable
vocabulary fully intact. Conversely, Llama and Qwen at 8 bits silence
hundreds of tokens while staying within three accuracy points and a few
perplexity points of full precision: the vocabulary shrinks before any
average moves much. Neither number predicts the other, so silencing is not a
restatement of perplexity---it is an independent property that a quality
average can hide in both directions.
The same holds language by language. We measured accuracy on Belebele
\citep{bandarkar2024belebele}, a reading-comprehension benchmark, in ten
languages for Qwen2.5-0.5B with its head quantized to 8 bits. Although
this head silences 550 of the 2550 Thai tokens, Thai accuracy does not
drop (32.1\% in full precision, 35.3\% quantized;
Table~\ref{tab:belebele}), and in no
language does the change follow the number of silenced tokens.

\begin{table}[t]
\centering\small
\setlength{\tabcolsep}{3pt}
\begin{tabular}{lrrrr}
\toprule
Language & FP & 8-bit & $\Delta$ & Silenced / script \\
\midrule
English  & 46.1 & 47.0 & $+0.9$ &  12 / 94{,}739 \\
Chinese  & 45.1 & 47.3 & $+2.2$ &  89 / 25{,}514 \\
Russian  & 38.2 & 39.2 & $+1.0$ &   0 / 4{,}149 \\
Korean   & 35.2 & 36.2 & $+1.0$ &   0 / 3{,}594 \\
Japanese & 34.3 & 35.3 & $+1.0$ &   0 / 1{,}471 \\
Thai     & 32.1 & 35.3 & $+3.2$ & 550 / 2{,}550 \\
Arabic   & 30.4 & 30.6 & $+0.1$ &  22 / 3{,}974 \\
Greek    & 29.0 & 30.7 & $+1.7$ &   3 / 233 \\
Hindi    & 28.0 & 29.6 & $+1.6$ &   0 / 56 \\
Hebrew   & 27.9 & 28.6 & $+0.7$ &  12 / 3{,}183 \\
\bottomrule
\end{tabular}
\caption{Belebele accuracy (\%) of Qwen2.5-0.5B in full precision and
with its head quantized to 8 bits, next to the silenced tokens of each
language's script. With 900 questions per language the standard error is
about 1.6 points; chance level is 25\%.}
\label{tab:belebele}
\end{table}

\begin{table}[t]
\centering\small
\setlength{\tabcolsep}{4pt}
\begin{tabular}{llrrr}
\toprule
Model & Bits & PPL & ARC (\%) & Silenced \\
\midrule
GPT-2 & fp32 & $51.3$ & 40.4 & 0 \\
 & int8 & $60.2$ & 38.8 & $+0$ \\
 & int4 & $3.2\!\times\!10^{5}$ & 28.2 & $+0$ \\
 & int3 & $4.2\!\times\!10^{16}$ & 25.4 & $+0$ \\
\midrule
Llama-3.2-1B & fp32 & $19.1$ & 58.2 & 0 \\
 & int8 & $20.8$ & 55.3 & $+264$ \\
 & int4 & $68.4$ & 39.6 & $+290$ \\
 & int3 & $1.7\!\times\!10^{4}$ & 28.1 & $+322$ \\
\midrule
Qwen2.5-0.5B & fp32 & $21.1$ & 56.3 & 1401 \\
 & int8 & $23.5$ & 55.8 & $+834$ \\
 & int4 & $58.6$ & 47.0 & $+837$ \\
 & int3 & $3.9\!\times\!10^{5}$ & 24.0 & $+844$ \\
\bottomrule
\end{tabular}
\caption{Quality and silenced-token counts (identity head, one
representative scheme per bitwidth: int8 symmetric per-tensor, int4/3
symmetric per-channel). ``Silenced'' is the count relative to full
precision ($+N$); at 3 bits GPT-2 silences tokens only with per-tensor scales (175 under
symmetric RTN, all proven by exact collision, 70 of them numeric; Section~\ref{sec:silence}). Qwen's full-precision
count (1401) is an upper bound: only 4 of these rows are exact duplicates,
the rest form the cluster of under-trained rows described in
Section~\ref{sec:method}. Qwen's counts include padding rows of the head
beyond the tokenizer's vocabulary, which no text can produce: 141 of the
1401 in full precision and 130 of each quantized count. PPL is WikiText-2; ARC is ARC-Easy.}
\label{tab:main}
\end{table}

\section{The loss is language-skewed}\label{sec:lang}

\begin{figure*}[t]
  \centering
  \includegraphics[width=\textwidth]{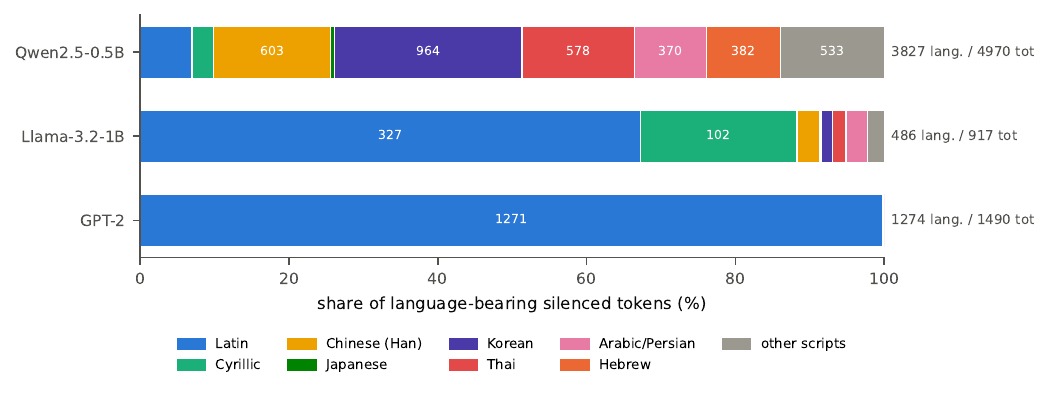}
  \caption{Language composition of the silenced tokens (union over
  quantized configurations, normalized to the language-bearing subset).
  Row labels give the language-bearing count and the total. The
  English-centric models lose mostly Latin tokens; the multilingual
  model loses overwhelmingly non-Latin ones.}
  \label{fig:lang}
\end{figure*}

The most striking pattern appears only where the vocabulary is
multilingual. We classify every silenced token by the Unicode script of its
characters---a declared proxy for language---and take, per model, the union
of distinct silenced tokens across all quantized configurations
(Figure~\ref{fig:lang}).

On the two English-centric models the language-bearing losses are mostly
Latin: GPT-2 loses $1271$ Latin tokens out of $1274$
language-bearing ones, and Llama-3.2-1B $327$ out of $486$; the
non-Latin remainder on Llama is mostly Cyrillic ($102$ tokens) with a long
thin tail of other scripts. On the multilingual Qwen vocabulary the picture
inverts: of $3827$ language-bearing silenced tokens only $266$ are Latin,
while the largest groups are Korean ($964$), Chinese ($603$), Thai ($578$),
Hebrew ($382$) and Arabic/Persian ($370$). The same model that loses barely
any English tokens loses hundreds in each of several other scripts. All of Qwen's and Llama's silenced tokens
are proven by exact collision in at least one configuration. On GPT-2
only $58$ of the $1274$ are; the rest are flagged tokens from its collapsed
2-bit per-channel head.
The pattern is not limited to the smallest model (Table~\ref{tab:scale}):
at 8 bits Qwen2.5-7B, like Qwen2.5-0.5B, loses mostly Thai tokens and
almost no Latin ones, while the Llama models lose almost only reserved
special tokens, from 1B to 70B. Qwen2.5-1.5B silences none: its
full-precision head already has 2233 silenced rows, 2212 of them exact
duplicates, and 8-bit rounding adds nothing. In
other words, quantization shrinks what the model can say most on exactly
the scripts a multilingual model is built to serve.

The script label is only a proxy: Cyrillic spans several languages, and the
Han ideographs are shared between Chinese and Japanese, so we keep Japanese
kana and Korean Hangul separate. Tokens that decode only to byte fragments,
and pure symbol or code fragments, are excluded from the language breakdown
and reported only in the totals. Even with these caveats the asymmetry is
large and consistent, and it is invisible to a single perplexity number.

\begin{table}[t]
\centering\small
\setlength{\tabcolsep}{3pt}
\begin{tabular}{lrrrrrr}
\toprule
 & & \multicolumn{2}{c}{Llama} & \multicolumn{3}{c}{Qwen2.5} \\
\cmidrule(lr){3-4}\cmidrule(lr){5-7}
 & GPT-2 & 1B & 70B & 0.5B & 1.5B & 7B \\
\midrule
Total    & 0 & 264 & 287 & 834 & 0 & 298 \\
\midrule
Latin    & 0 &   9 &   9 &  12 & 0 &   5 \\
Cyrillic & 0 &   5 &   5 &   0 & 0 &   0 \\
Han      & 0 &   0 &   0 &  89 & 0 &  26 \\
Thai     & 0 &   0 &   0 & 550 & 0 & 196 \\
Arabic   & 0 &   0 &   0 &  22 & 0 &  12 \\
Hebrew   & 0 &   0 &   0 &  12 & 0 &   5 \\
Special  & 0 & 235 & 254 &   8 & 0 &   5 \\
Other    & 0 &  15 &  19 &  11 & 0 &   7 \\
Padding  & 0 &   0 &   0 & 130 & 0 &  42 \\
\end{tabular}
\caption{Tokens newly silenced at 8 bits (symmetric per-tensor RTN), by
script and model size; all are exact collisions. Special: reserved and
control tokens; Padding: padding rows beyond the tokenizer's vocabulary. No
Korean or Japanese token is silenced at 8 bits.}
\label{tab:scale}
\end{table}

\paragraph{Why non-Latin tokens.} The pattern has a natural explanation:
under-trained tokens \citep{land2024magikarp}. A multilingual tokenizer
spends most of its vocabulary on the dominant script (here Latin), while
tokens that are rare in training keep short rows that sit very close
together, near the interior of the hull. In Qwen's full-precision head
about 1400 such rows already form a tight cluster. At 8 bits, rounding
merges the rows next to that cluster with it: every token newly silenced
by Qwen2.5-0.5B at 8 bits has a row of norm about $0.30$, the same as the
cluster, and ends up identical to cluster rows. These under-trained tokens
are not spread evenly across scripts. Over the whole vocabulary, the
median row norm is $0.30$ for Thai against $0.46$ overall, while Korean
($0.42$), Hebrew ($0.40$), Arabic ($0.45$) and Chinese ($0.47$) are close to
the global value. Thai is therefore under-trained as a whole, whereas in
the other scripts only a tail of tokens is. Relative to vocabulary size
the loss is still large: across all configurations Qwen loses $27\%$ of
its Korean tokens, $23\%$ of its Thai tokens and $12\%$ of its Hebrew
tokens, but only $0.3\%$ of its Latin tokens.

This mechanism is measurable. The median row norm of the silenced tokens
sits far below that of the rest: $0.31$ versus $0.46$ on Qwen2.5-0.5B and
$0.55$ versus $0.99$ on Llama-3.2-1B, and low row norm alone almost
perfectly separates silenced from safe tokens (AUC $0.99$ and $1.00$,
respectively, for predicting silencing from the norm). GPT-2 is the
instructive exception (AUC $0.63$): there silencing is driven not by
rarity but by the exact grid collisions of the digit rows at 3 bits.

\section{When does it matter?}\label{sec:matter}

Silencing a deep-tail reserved token that no input ever needs is harmless;
silencing a token the model must produce is not. The clearest harmful case
is GPT-2 at 3 bits. There, digit rows collide exactly, so the digits are
proven silenced. On 400 WikiText-103 documents we find $52$ positions where the
full-precision model greedily decodes the correct digit ($120$ including multi-digit
numeric tokens such as years) but the quantized
model cannot produce it at all---the outcome is impossible by construction,
before any other degradation is considered.

This shows that silencing is
not a purely theoretical curiosity: when it lands on task-relevant tokens,
it removes correct outputs outright. Whether it matters therefore depends
on \emph{which} tokens are silenced---exactly the composition that
Figures~\ref{fig:cats} and~\ref{fig:lang} make visible.

\begin{table}[t]
\centering\small
\setlength{\tabcolsep}{3pt}
\begin{tabular}{lrrrr}
\toprule
Script & Silenced & As id & As word & Inside word \\
\midrule
Thai           & 550 & 0 & --          & 545 (465{,}566) \\
Chinese        &  89 & 0 & --          & 0 \\
Arabic/Persian &  22 & 0 & 6 (122)     & 22 (6{,}386) \\
Hebrew         &  12 & 0 & 2 (3)       & 2 (975) \\
Latin          &  12 & 0 & 0           & 1 (15) \\
\bottomrule
\end{tabular}
\caption{Occurrences in Wikipedia text (34 languages, 2000 articles each)
of the tokens Qwen2.5-0.5B silences at 8 bits (symmetric per-tensor):
number of tokens found, with total occurrences in parentheses. On this text
the tokenizer never produces a silenced token; most Thai and all Arabic ones
occur inside longer words. ``--'': no word boundaries in the script.}
\label{tab:corpus}
\end{table}

\paragraph{What happens at realistic bit-widths.} The digit failure needs aggressive 3-bit quantization. At the more common 8-bit setting, the dynamics differ but reveal latent structural risks: while GPT-2 silences nothing, Llama-3.2-1B and Qwen2.5-0.5B already silence hundreds of tokens. Inspecting those tokens, they are almost entirely deep-tail: byte fragments, reserved special tokens, under-trained ``glitch'' tokens, and rare non-Latin subword pieces---not common words. Such tokens seldom top the distribution on
ordinary text, which is why 8-bit silencing rarely shows up in a greedy
generation and why our corpus check below finds no error. To measure how often these tokens occur in real text, we counted them in
Wikipedia articles in 34 languages (Table~\ref{tab:corpus}). On this text
the tokenizer never produces any of the silenced tokens: it encodes the same
strings with other tokens. As strings they do occur, mostly inside longer
words: the Thai ones about half a million times. On Llama-3.2-1B, 235 of
the 264 tokens silenced at 8 bits are reserved special tokens, and none of
the 264 appears in the tokenized text. Over 32 million positions of
Wikipedia text, a silenced token is never the correct next token for
Llama-3.2-1B, Qwen2.5-0.5B or Qwen2.5-7B at 8 bits.
A targeted test of prompts whose correct continuation is a silenced token
therefore finds no prompts to test in this text at 8 bits; where such prompts exist, as
for GPT-2 at 3 bits above, the quantized model fails on all of them. This makes 8-bit
silencing a \emph{latent} risk: it quietly removes rare and special tokens
(including control tokens and low-resource subwords) that a targeted input
could still call for, while leaving common-token behaviour---and average
metrics---largely untouched.

The converse also holds, confirming that structural reachability and empirical corpus evaluation remain complementary. Running the same check for Qwen at 4 bits over Chinese text (mixed
English--Chinese Wikipedia corpus), we find
\emph{no} position where a silenced token was the correct full-precision
prediction: the silenced non-Latin tokens here are deep-tail types this
particular corpus never required. A structural loss need not produce a
measurable error on a given corpus. This is exactly why silencing should be
reported as its own quantity: it is neither always harmful nor always
harmless, and averages cannot tell the two cases apart.

\section{Related work}

\paragraph{Output-layer geometry.} Unargmaxable tokens were introduced by
\citet{grivas2022} as a property of trained output heads: rows inside the
convex hull of the others can never win the argmax. This is the structural
face of the stolen-probability effect, in which rare tokens are crowded out
of the probability simplex \citep{demeter2020}, and it is enabled by the
low-rank softmax bottleneck \citep{yang2018softmax}, the same structure that
lets one recover hidden geometry through a production API
\citep{finlayson2024}. This line of work studies full-precision models; we
ask how quantization reshapes the set of reachable tokens.

\paragraph{Quantization.} Weight-only post-training quantization ranges from
error-correcting and activation-aware rounding
\citep{frantar2023gptq,lin2024awq} to calibrated schemes
\citep{shao2024omniquant} and equivalence-preserving transforms such as
smoothing and learned rotations
\citep{xiao2023smoothquant,ashkboos2024quarot,liu2024spinquant}. All of them
optimize average reconstruction error or downstream loss; none targets
reachability, and the rounding step they share is precisely what silences
tokens. Deployment kernels \citep{dettmers2022llmint8,dettmers2023qlora}
often leave the head in higher precision and so avoid structural silencing,
at the cost of a larger model.

\paragraph{Fidelity and reachability.} Token-level fidelity metrics compare
the outputs of a quantized and a full-precision model
\citep{displacement2026}, and lexical-reachability scores ask which words a
decoder can produce under a given sampling filter \citep{wcs2026}. Both are
complementary to our question: the former compares distributions on a
corpus and the latter concerns the decoding procedure, whereas we
characterize, independently of any corpus or decoder, which tokens the
model \emph{could} ever produce. To our knowledge this structural,
language-resolved view of quantization is new.

\section{Discussion}

Our findings yield three practical implications. First, silencing is \emph{cheap to measure}: the estimator runs on the output head alone, in minutes to hours
on a single GPU, and needs no data. It can be run as a routine check
whenever a model is quantized, next to the usual perplexity pass. Second,
the right thing to report is not a single number but a \emph{range with a
script breakdown}: the proven-silenced count (exact collisions) as a lower
end, that count plus the flagged tokens as an upper end, and the per-language composition
that tells a practitioner whether the loss lands on tokens they care about.
Third, our stress test uses plain RTN, but the mechanism---rounding rows
onto a shared grid until marginal ones fall inside the hull---is generic;
stronger quantizers change how quickly it happens, not whether it can.

Quantizing a model to 8 bits is often assumed to be lossless. For
practitioners, the main point is to check the precision of the model's
output head. Quantizing the head to 8 bits already silences hundreds of
tokens on large-vocabulary models: reserved tokens on Llama, mostly
non-Latin tokens on Qwen. Whether
this happens depends on the deployment format: bitsandbytes and the public
GPTQ checkpoints we inspected keep the head in 16 bits, whereas
llama.cpp's GGUF formats quantize it (for example to Q6\_K). Keeping the
head in higher precision prevents new silencing, but not the loss of
tokens that a full-precision head already cannot emit.

The clearest recommendation is for multilingual deployment. A 4-bit model
that keeps its English scores can still have silenced a meaningful slice of
a low-resource script, and no English-only evaluation will reveal it. A script-level silenced-token count is a small, direct safeguard against deploying a model that has quietly lost the ability to generate text in some of the languages it was trained to serve.

\section{Conclusion}

Quantization can make tokens impossible to produce. We gave a simple,
scalable way to estimate how many tokens a quantized output head silences,
with a clear meaning for each classification, and showed that silencing appears
already at 8 bits on large-vocabulary models and falls disproportionately
on non-Latin scripts. Because a single perplexity or accuracy number hides
this, we argue that silenced-token counts---and especially their
language breakdown---should be reported alongside standard metrics when
quantizing multilingual models.

\section*{Limitations}

Our classifications are relative to the stored weight matrix; on clusters of
near-duplicate rows the flagged/safe classification is intrinsically
unstable, and we report margin bounds rather than a binary answer there.
The guarantee concerns greedy decoding: a silenced token keeps a nonzero
probability and can still be emitted under sampling.
The analysis is of the output head alone: it is exact for whether a token
\emph{can} win, but it is an upper bound on what the full network actually
realizes on real inputs, and deployment pipelines differ in whether they
quantize the head at all. Besides RTN, we test GPTQ and AWQ on four models only; smoothing,
rotation-based and other quantizers \citep{xiao2023smoothquant,
ashkboos2024quarot,liu2024spinquant,zhang2024qqq} remain untested. The mapping from
Unicode script to language is a proxy, not a language identifier, and our
multilingual evidence rests on a single multilingual vocabulary, shared
by the three Qwen2.5 models; the
pattern remains to be confirmed on other multilingual heads. Not every count is exact: flagged tokens remain in collapsed 2-bit heads,
in Qwen's full-precision cluster and in the GPTQ head of Qwen2.5-7B.
Finally, our evidence on task impact is limited to corpus checks and to
Belebele on a 0.5B model; a full downstream evaluation is left to future
work.

\bibliography{refs}

\end{document}